\documentclass{article} 
\usepackage{subfigure} 
\usepackage{amsmath,amsthm,amssymb}
\usepackage{hyperref}
\usepackage[pdftex]{graphicx}

\usepackage{natbib}
\setcitestyle{numbers,square}
\usepackage{algorithm}
\usepackage{algorithmic}
\usepackage{fancyvrb}
\usepackage{verbatim}
\usepackage{wrapfig}
\usepackage{color}
\usepackage{stfloats}
\usepackage[accepted]{icml2013}

\newcommand{\x}{\mathbf{x}}

\newcommand{\bl}{\boldsymbol{\beta}}

\newcommand{\Fb}{\mathbf{F}}
\newcommand{\Eddie}{\textcolor{black}}
\newcommand{\Matt}{\textcolor{black}}

\icmltitlerunning{Cost-Sensitive Tree of Classifiers}

\begin{document}

\twocolumn[
\icmltitle{Cost-Sensitive Tree of Classifiers}

\icmlauthor{Zhixiang (Eddie) Xu}{xuzx@cse.wustl.edu}
\icmlauthor{Matt J. Kusner}{mkusner@wustl.edu}
\icmlauthor{Kilian Q. Weinberger}{kilian@wustl.edu}
\icmlauthor{Minmin Chen}{mchen@wustl.edu}
\icmladdress{Washington University,
            One Brookings Dr., St. Louis, MO 63130 USA}

\icmlkeywords{cost-sensitive, budgeted learning, tree structure,  machine learning, ICML}

\vskip 0.3in
]

\begin{abstract}
Recently, machine learning algorithms have successfully entered large-scale real-world industrial applications (\emph{e.g.} search engines and email spam filters). Here, the CPU cost during test-time must be budgeted and accounted for. 
In this paper, we address the challenge of balancing the test-time cost and the classifier accuracy in a principled fashion. The test-time cost of a classifier is often dominated by the computation required for feature extraction---which can vary drastically across features. 
We decrease this extraction time
by constructing a tree of classifiers, through which test inputs traverse along individual paths. Each path extracts different features and is optimized for a specific sub-partition of the input space. 
By only computing features for inputs that benefit from them the most, our cost-sensitive tree of classifiers can match the high accuracies of the current state-of-the-art at a small fraction of the computational cost. 
\end{abstract}

\section{Introduction}
Machine learning algorithms are widely used in many real-world applications, ranging from email-spam~\citep{weinberger2009feature} and adult content filtering~\citep{fleck1996finding}, to web-search engines~\citep{zheng2007general}. As machine learning transitions into these industry fields, managing the CPU cost at test-time becomes increasingly important. In applications of such large scale, computation must be budgeted and accounted for. Moreover, reducing energy wasted on unnecessary computation can lead to monetary savings and reductions of greenhouse gas emissions.

The \emph{test-time cost} consists of the time required to evaluate a classifier and the time to extract features for that classifier, where the extraction time across features is highly variable. 
Imagine introducing a new feature to an email spam filtering algorithm that requires $0.01$ seconds to extract per incoming email. If a web-service receives one billion emails (which many do daily), it would require 115 extra CPU days to extract just this feature. Although this additional feature may increase the accuracy of the filter, the cost of computing it for \emph{every email} is prohibitive. This introduces the problem of balancing the test-time cost and the classifier accuracy.  
Addressing this trade-off in a principled manner is crucial for the applicability of machine learning.

In this paper, we propose a novel algorithm, \emph{Cost-Sensitive Tree of Classifiers} (CSTC). 
A CSTC tree  (illustrated schematically in Fig.~\ref{fig:tronus}) is a tree of classifiers that is carefully constructed to reduce the \emph{average} test-time complexity of machine learning algorithms, while maximizing their accuracy. Different from prior work, which reduces the total cost for every input~\cite{efron2004least} or which stages the feature extraction into linear cascades~\cite{viola2004robust,Lefakis2010,Saberian2010,pujara2011using,chen2011}, a CSTC tree incorporates \emph{input-dependent feature selection} into training 
and dynamically allocates  higher feature budgets for infrequently traveled tree-paths. By introducing a probabilistic tree-traversal framework, we can compute the exact expected test-time cost of a CSTC tree. CSTC is trained with a single global loss function, whose test-time cost penalty is a direct relaxation of this expected cost. This principled approach leads to unmatched  test-cost/accuracy tradeoffs as it naturally divides the input space into sub-regions and extracts expensive features only when necessary. 




We make several novel contributions: 
1. We introduce the meta-learning framework of CSTC trees and derive the expected cost of an input traversing the tree during test-time. 
2. We relax this expected cost with a mixed-norm relaxation and derive a single global optimization problem to train all classifiers jointly. 
3. We demonstrate on synthetic data that CSTC effectively allocates features to classifiers where they are most beneficial and show on large-scale real-world web-search ranking data that CSTC significantly outperforms the current state-of-the-art in test-time cost-sensitive learning---maintaining the performance of the best algorithms for web-search ranking at a fraction of their computational cost.

\section{Related Work}

A basic approach to control test-time cost  is the use of $l_1$-norm regularization~\citep{efron2004least}, 
which results in a sparse feature set, and can significantly reduce the feature cost during test-time (as unused features are never computed). 
However, this approach fails to address the fact that some inputs may be successfully classified by only a few cheap 
features, whereas others strictly require expensive features for correct classification. 

There is much previous work that extends single classifiers  
to classifier cascades (mostly for binary classification) 
\citep{viola2004robust,Lefakis2010,Saberian2010,pujara2011using,chen2011}. 
In these cascades, 
several classifiers are ordered into a sequence of stages.
Each classifier can either reject  
inputs (predicting them), 
or pass them on to the next stage, based on the prediction of each input. 
To reduce the test-time cost, these cascade algorithms enforce that classifiers in early stages use very few and/or cheap features and reject many easily-classified inputs. Classifiers in later stages, however, are more expensive and cope with more difficult inputs. 
This linear structure is particularly effective for applications with highly skewed class imbalance and generic features. One celebrated example is face detection in images, where the majority of all image regions do not contain faces and can often be easily rejected based on the response of a few simple Haar features~\citep{viola2004robust}. 
The linear cascade model is however less suited for learning tasks with \emph{balanced classes} and \emph{specialized features}. It cannot fully capture the scenario where different partitions of the input space require different expert features, as all inputs follow the same linear chain. 


\citet{grubbspeedboost} and \citet{greedymiser} focus on training a classifier that explicitly trades-off test-time cost and accuracy. Instead of optimizing the trade-off by building a cascade, they push the cost trade-off into the construction of the weak learners. It should be noted that, in spite of the high accuracy achieved by these techniques, the algorithms are based heavily on stage-wise regression (gradient boosting) \citep{friedman2001greedy}, and are less likely to work with more general weak learners. 

\citet{GaoKoller11} use locally weighted regression during test time to predict the information gain of unknown features. 
Different from our algorithm, their model is learned during test-time, which introduces an additional cost especially for large data sets. In contrast, our algorithm learns and fixes a tree structure in training and has a test-time complexity that is constant with respect to the training set size. 

\citet{karayev2012timely} use reinforcement learning to dynamically select features to maximize the average precision over time in an object detection setting. In this case, the dataset has multi-labeled inputs and thus warrants a different approach than ours.

Hierarchical Mixture of Experts (HME)~\citep{jordan1994hierarchical} also builds tree-structured classifiers. However, in contrast to CSTC, this work is not motivated by reductions in test-time cost and results in fundamentally different models. In CSTC, each classifier is trained with the test-time cost in mind and each test-
input only traverses a \emph{single} path from the root down to a terminal element, accumulating path-specific costs. In HME, 
all test-inputs traverse all paths and all leaf-classifiers  contribute to the final prediction, incurring the same cost for all test-inputs. 

Recent tree-structured classifiers include the work of \citet{deng2011fast}, who speed up the training and evaluation of label trees \citep{bengio2010label}, by avoiding many binary one-vs-all classifier evaluations. Differently, we focus on problems in which feature extraction time dominates the test-time cost which motivates different algorithmic setups. \citet{dredze2007learning} combine the cost to select a feature with the mutual information of that feature to build a decision tree that reduces the feature extraction cost. Different from this work, they do not directly minimize the total test-time cost of the decision tree or the risk. Possibly most similar to our work are \cite{busa2012fast}, who learn a directed acyclic graph via a Markov decision process to select features for different instances, and \cite{wang2012local}, who adaptively partition the feature space and learn local region-specific classifiers. Although each work is similar in motivation, the algorithmic frameworks are very different and can be regarded complementary to ours.

\section{Cost-sensitive classification}
\label{sec:method}
We first introduce our notation and then formalize our test-time cost-sensitive learning setting.
%
Let the training data consist of inputs $\mathcal{D}\!=\!\{ \x_1,\dots,\x_n\}\subset\mathcal{R}^d$ with corresponding class labels $\{ y_1,\dots,y_n\}\subseteq\mathcal{Y}$, where $\mathcal{Y}=\mathcal{R}$ in the case of regression ($\mathcal{Y}$ could also be a finite set of categorical labels---because of space limitations we do not focus on this case in this paper). 

\textbf{Non-linear feature space.}
Throughout this paper, we focus on linear classifiers but in order to allow non-linear decision boundaries we map the input into a non-linear feature space with the ``boosting trick''~\citep{friedman2001greedy,chapelle2011boosted}, prior to our optimization. In particular, we first train gradient boosted regression trees with a squared loss penalty~\citep{friedman2001greedy}, $H'(\x_i)\!=\!\sum_{t=1}^T h_t(\x_i)$, where each function $h_t(\cdot)$ is a limited-depth CART tree~\citep{breiman1984classification}. We then apply the mapping $\x_i\!\rightarrow\! \phi(\x_i)$ to all inputs, where $\phi(\x_i)= [h_1(\x_i), \dots, h_T(\x_i)]^\top$. 
To avoid confusion between CART trees and the CSTC tree, we refer to CART trees $h_t(\cdot)$ as \emph{weak learners}. 

\textbf{Risk minimization.} At each node in the CSTC tree we propose to learn a linear classifier in this feature space, $H(\x_i) = \phi(\x_i)^\top\bl$ with $\bl\in\mathcal{R}^T$, which is trained to explicitly reduce the CPU cost during test-time.
We learn the weight-vector $\bl$ by minimizing a convex empirical risk function $\ell(\phi(\x_i)^\top\bl,y_i)$ with $l_1$ regularization, $|\bl|$. In addition, we incorporate a cost term $c(\bl)$, which we derive in the following subsection, to restrict test-time cost. The combined test-time cost-sensitive loss function becomes
\begin{align}
	{\cal L}(\bl) = \Eddie{\underbrace{\sum_i\ell(\phi(\x_i)^\top\bl,y_i) + \rho |\bl|}_{\textrm{regularized risk}}} + \lambda\underbrace{c(\bl)}_{\textrm{test-cost}},	\label{eq:onenodeloss}
\end{align}
where $\lambda$ is the accuracy/cost trade-off parameter, and $\rho$ controls the strength of the regularization. 

\textbf{Test-time cost.} 
There are two factors that contribute to the test-time cost of each classifier. The weak learner evaluation cost of all active $h_t(\cdot)$ (with $|\beta_t|\!>\!0$) and the feature extraction cost for all features used in these weak learners. 
We assume that features are computed \emph{on demand} with the cost $\mathbf{c}$ the first time they are used, and are free for future use (as feature values can be cached).
We define an auxiliary matrix $\Fb\!\in\!\{ 0,1\}^{d \times T}$ with $F_{\alpha t}\!=\!1$ if and only if the weak learner $h_t$ uses feature $f_\alpha$. 
Let $e_t\!>\!0$ be the cost to evaluate a $h_t(\cdot)$, and $c_\alpha$ be the cost to extract feature $f_\alpha$. With this notation, we can formulate the total test-time cost for an instance precisely as 
\begin{align}
	c(\bl) = \underbrace{\sum_{t} e_{t}\|\beta_t\|_0}_{\textrm{evaluation cost}} + \underbrace{\sum_{\alpha} c_\alpha \left\|\sum_{t} |F_{\alpha t}\beta_t|\right\|_0}_{\textrm{feature extraction cost}}, \label{eq:noncont}
\end{align}
where the $l_0$ norm for scalars is defined as $\|a\|_0 \!\in\! \{ 0,1\}$ with $\|a\|_0\!=\!1$ if and only if $a\!\neq\!0$. 
The first term assigns cost $e_t$ to every \Eddie{weak learner} used in $\bl$, the second term assigns cost $c_{\alpha}$ to every feature that is extracted by \emph{at least one} of such weak learners. 

\textbf{Test-cost relaxation.}
\label{subsec:singlenode}
The cost formulation in (\ref{eq:noncont}) is exact but difficult to optimize as the $l_0$ norms are non-continuous and non-differentiable. As a solution, throughout this paper we use the mixed-norm relaxation of the $l_0$ norm over sums,
\begin{align}
\sum_{j} \left\| \sum_{i} |a_{ij}| \right\|_0 \rightarrow \sum_j\sqrt{\sum_{i} (a_{ij})^2}, \label{eq:l0relaxation}
\end{align}
described by \citep{kowalski2009sparse}. Note that for a single element this relaxation relaxes the $l_0$ norm to the $l_1$ norm, $\| a_{ij} \|_0\! \rightarrow\! \sqrt{(a_{ij})^2}\! =\! |a_{ij}|$, and recovers the commonly used approximation to encourage sparsity~\citep{efron2004least,scholkopf2001learning}. We plug the cost-term (\ref{eq:noncont}) into the loss in~(\ref{eq:onenodeloss}) and apply the relaxation (\ref{eq:l0relaxation}) to all $l_0$ norms to obtain
\begin{align}
\underbrace{\sum_i\!\ell_i \!+\! \rho |\bl|}_{\textrm{regularized loss}} + \lambda \Bigg(\!\!\!\!\underbrace{\sum_{t}\! e_t |\beta_t|}_{\textrm{ev. cost penalty}} \!+\! \underbrace{\sum_{\alpha} \!c_\alpha\! \sqrt{\sum_{t}(F_{\alpha t}\beta_t)^2}}_{\textrm{feature cost penalty}}\!\Bigg), \label{eq:onenode}
\end{align}
where we abbreviate $\ell_i\!=\!\ell(\phi(\x_i)^\top\bl,y_i)$ for simplicity.
While (\ref{eq:onenode}) is cost-sensitive, it is restricted to a single linear classifier. In the next section we describe how to expand this formulation into a cost-effective tree-structured model. 

\section{Cost-sensitive tree}
\label{method:tronus_tree}
\begin{figure}[t]
\centerline{
\centerline{\includegraphics[width = 0.9618\columnwidth]{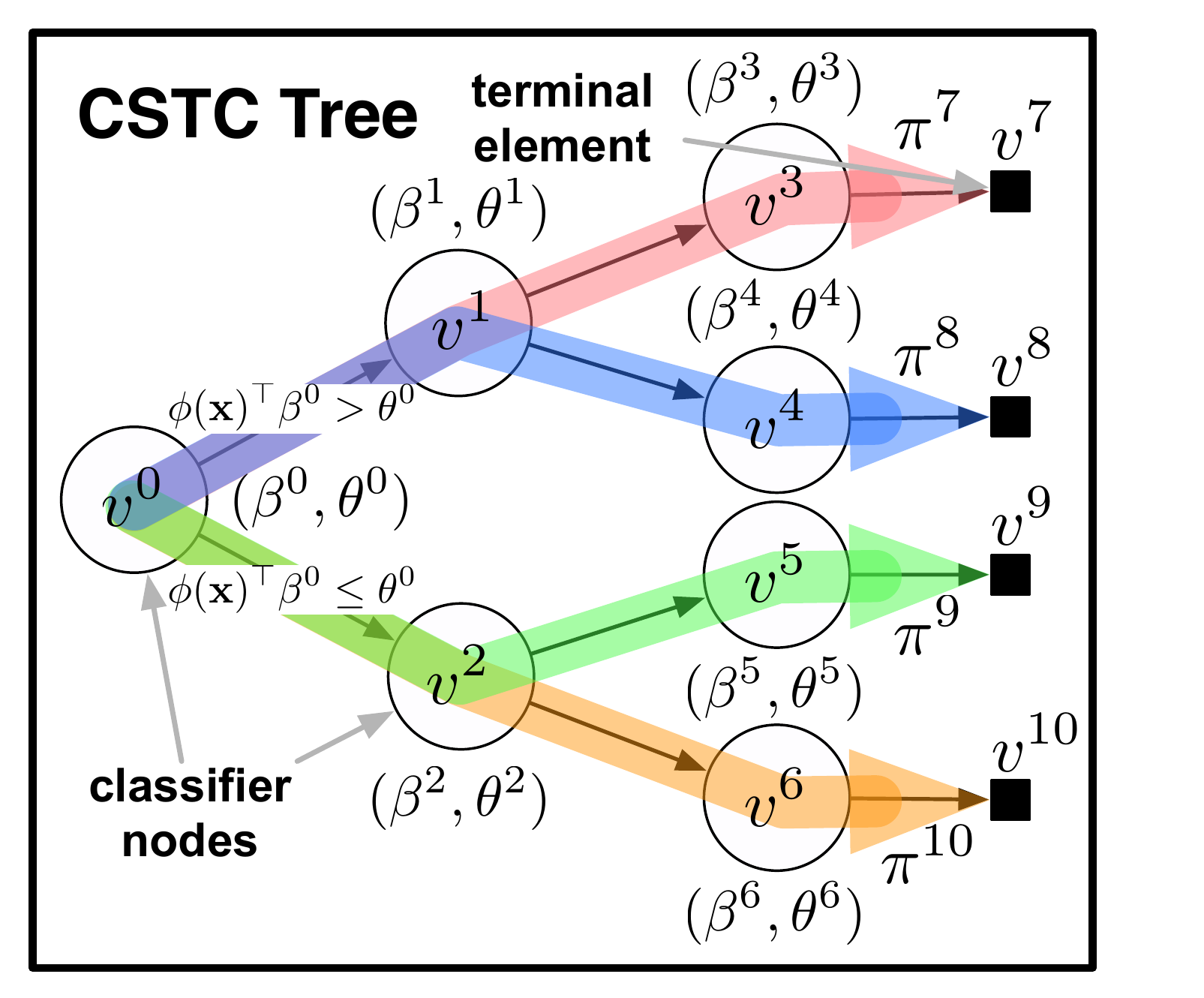}}
}
\caption{A schematic layout of a CSTC tree. Each node $v^k$ has a threshold $\theta^k$ to send instances to different parts of the tree and a weight vector $\bl^k$ for prediction. We solve for $\bl^k$ and $\theta^k$ that best balance the accuracy/cost trade-off for the whole tree. All paths of a CSTC tree are shown in color.} 
\label{fig:tronus}
\end{figure}

\begin{table*}[t!]	
\begin{align}
	\min_{\bl^1,\theta^1,\dots,\Eddie{\bl^{|V|},\theta^{|V|}}}\sum_{v^k\in V}\! \underbrace{\left(\frac{1}{n}\!\sum_{i=1}^n p_i^k\ell_i^k \!+
	\!\rho |\bl^k|\! \right)}_{\textrm{regularized risk}} +
	\lambda \!
	\sum_{v^l\in L}p^l\!\Bigg[\underbrace{\sum_t \!e_t\! \sqrt{\!\sum_{v^j \in \pi^l}\!(\beta_t^j)^2} }_{\textrm{evaluation cost penalty}}
	\!+\! 
	\underbrace{\sum_\alpha \!c_\alpha\! \sqrt{ \sum_{v^j \in \pi^l} \sum_t (F_{\alpha t} \beta_t^j)^2}}_{\textrm{feature cost penalty}}\Bigg]\label{eq:nodeloss}
\end{align}
\vspace{-5.5ex}
\end{table*}
We begin by introducing foundational concepts regarding the CSTC tree and derive a global loss function~(\ref{eq:nodeloss}). Similar to the previous section, we first derive the exact cost term and then relax it with the mixed-norm. Finally, we describe how to optimize this function efficiently and to undo some of the inaccuracy induced by the mixed-norm relaxations. 

\textbf{CSTC nodes.}  
We make the assumption that instances with similar labels can utilize similar features.\footnote{For example, in web-search ranking, features generated by browser statistics are typically predictive only for highly relevant pages as they require the user to spend significant time on the page and interact with it.}
We therefore design our tree algorithm to partition the input space based on classifier predictions. Classifiers that reside deep in the tree become experts for a small subset of the input space and intermediate classifiers determine the path of instances through the tree. 
We distinguish between two different elements in a CSTC tree (depicted in Figure \ref{fig:tronus}): \emph{classifier nodes} (white circles) and \emph{terminal elements} (black squares). Each \emph{classifier node} $v^k$ is associated with a weight vector $\bl^k$ and a threshold $\theta^k$. Different from cascade approaches, these classifiers not only classify inputs using $\bl^k$, but also branch them by their threshold $\theta^k$, sending inputs to their upper child if $\phi(\x_i)^\top\bl^k\!>\!\theta^k$, and to their lower child otherwise.  
\emph{Terminal elements} are ``dummy" structures and are \emph{not} classifiers. They return the predictions of their direct parent classifier nodes---essentially functioning as a placeholder for an exit out of the tree. The tree structure may be a full balanced binary tree of some depth (eg. figure~\ref{fig:tronus}), or can be pruned based on a validation set (eg. figure~\ref{fig:jaccard}, left).

During test-time, inputs are first applied to the root node $v^0$. The root node produces predictions $\phi(\x_i)^\top \bl^{0}$ and sends the input $\x_i$ along one of two different paths, depending on whether $\phi(\x_i)^\top \bl^{0}\! >\! \theta^{0}$. By repeatedly branching the test-inputs, classifier nodes sitting deeper in the tree only handle a small subset of all inputs and become specialized towards that subset of the input space.

\subsection{Tree loss}
\label{sec:lossfunction}
We derive a single global loss function over all nodes in the CSTC tree. 

\textbf{Soft tree traversal.} 
Training the CSTC tree with hard thresholds leads to a combinatorial optimization problem, which is NP-hard. Therefore, during training, we \emph{softly} partition the inputs and assign \emph{traversal probabilities} \mbox{$p(v^k | \x_i)$} to denote the likelihood of input $\x_i$ traversing through node $v^k$. Every input $\x_i$ traverses through the root, so we define $p(v^0 | \x_i)\!=\!1$ for all $i$. We use the sigmoid function to define a soft belief that an input $\x_i$ will transition from classifier node $v^k$ to its \emph{upper} child $v^j$ as $ p(v^j | \x_i, v^k)\!=\!\sigma(\phi(\x_i)^{\top}\bl^k\!-\!\theta^k)$.
\footnote{The sigmoid function is defined as  $\sigma(a)\!=\!\frac{1}{1+exp(-a)}$ and takes advantage of the fact that $\sigma(a)\in[0,1]$ and that $\sigma(\cdot)$ is strictly monotonic.} 
The probability of reaching child $v^j$ from the root is, recursively, $p(v^j | \x_i) = p(v^j | \x_i, v^k)p(v^k | \x_i)$, because each node has exactly one parent. 
For a \emph{lower} child $v^l$ of parent $v^k$ we naturally obtain $p(v^l | \x_i) = \big[1 - p(v^j | \x_i, v^k)\big]p(v^k | \x_i)$. 
In the following paragraphs we incorporate this probabilistic framework into the single-node 
risk and cost terms of eq.~(\ref{eq:onenode}) to obtain the corresponding \emph{expected} tree risk and tree cost. 

\textbf{Expected tree risk.}
The \emph{expected tree risk} can be obtained byWg over all nodes $V$ and inputs and 
weighing the risk $\ell(\cdot)$ of input $\x_i$ at node $v^k$  by the probability \mbox{$p_i^k\! =\! p(v^k | \x_i)$}, 
\begin{align}
\frac{1}{n}	 \sum_{i=1}^n \sum_{v^k\in V} p_i^k\ell(\phi(\x_i)^\top\bl^k,y_i).\label{eq:weighted_loss}
\end{align}
This has two effects: 1. the local risk for each node focusses more on likely inputs; 
2. the global risk attributes more weight to classifiers that serve many inputs. 


\textbf{Expected tree costs.}
The cost of a test-input is the cumulative cost across all classifiers along its path through the CSTC tree. 
Figure~\ref{fig:tronus} illustrates an example of a CSTC tree with all paths highlighted in color. Every test-input must follow along exactly one of the paths from the root to a terminal element.
Let $L$ denote the set of all terminal elements (\emph{e.g.}, in figure~\ref{fig:tronus} we have $L\!=\!\{v^7,v^8,v^9,v^{10}\}$), and for any $v^l\!\in\!L$ let $\pi^l$ denote the set of all \emph{classifier nodes} along the unique path from the root $v^0$ before terminal element $v^l$ (\emph{e.g.}, $\pi^{9}\!=\!\{v^0,v^2,v^5\}$). 
The evaluation and feature cost of this unique path is exactly
\begin{align}
c^l\!=\!\underbrace{\sum_t  e_t \Bigg\| \sum_{v^j \in \pi^l} |\beta_t^j| \Bigg\|_0}_{\textrm{evaluation cost}} + \underbrace{\sum_\alpha c_\alpha \Bigg\|\sum_{v^j \in \pi^l} \sum_t |F_{\alpha t}\beta_t^j| \Bigg\|_0}_{\textrm{feature cost}}. \nonumber
\end{align}
This term is analogous to eq. (\ref{eq:noncont}), except the cost $e_t$ of the weak learner $h_t$ is paid if \emph{any} of the classifiers $v^j$ in path $\pi^l$ use this tree (\emph{i.e.} assign $\beta_t^j$ non-zero weight). Similarly, the cost $c_\alpha$ of a feature $f_\alpha$ is paid exactly once if any of the weak learners of any of the classifiers along $\pi^l$ require it. Once computed, a feature or weak learner can be reused by all classifiers along the path for free (as the computation can be cached very efficiently). 

Given an input $\x_i$, the probability of reaching terminal element $v^l \in L$ (traversing along path $\pi^l$) is $p_i^l = p(v^l | \x_i)$. Therefore, the marginal probability that a training input (picked uniformly at random from the training set) reaches $v^l$ is $p^l \!=\! \sum_ip(v^l | \x_i)p(\x_i) \!=\! \frac{1}{n}\sum_{i=1}^n p_i^l$.
 With this notation, the \emph{expected cost} for an input traversing the CSTC tree becomes $\mathbb{E}[c^l]=\sum_{v^l\in L}p^lc^l$. 
Using our $l_0$-norm relaxation in eq.~(\ref{eq:l0relaxation}) on both $l_0$ norms in $c^l$ gives the final expected tree cost penalty
\begin{align}
\sum_{v^l \in L}\! p^l \Bigg[ \!\!\sum_t  e_t \!\sqrt{ \sum_{v^j \in \pi^l} (\beta_t^j)^2 } + \!\!\sum_\alpha c_\alpha \!\sqrt{ \sum_{v^j \in \pi^l} \sum_t (F_{\alpha t}\beta_t^j)^2 } \Bigg],\nonumber
\end{align}
which naturally encourages weak learner and feature re-use along paths through the CSTC tree.

\textbf{Optimization problem.}
We combine the risk (\ref{eq:weighted_loss}) with the cost penalties and add the $l_1$-regularization term  (which is unaffected by our probabilistic splitting) to obtain the global optimization problem (\ref{eq:nodeloss}). (We abbreviate the empiWisk at node \Eddie{$v^k$} as $\ell_i^k\!=\!\ell(\phi(\x_i)^\top\bl^k,y_i)$.)

\begin{figure*}[t]
\centerline{
\includegraphics[width = 0.9618\textwidth]{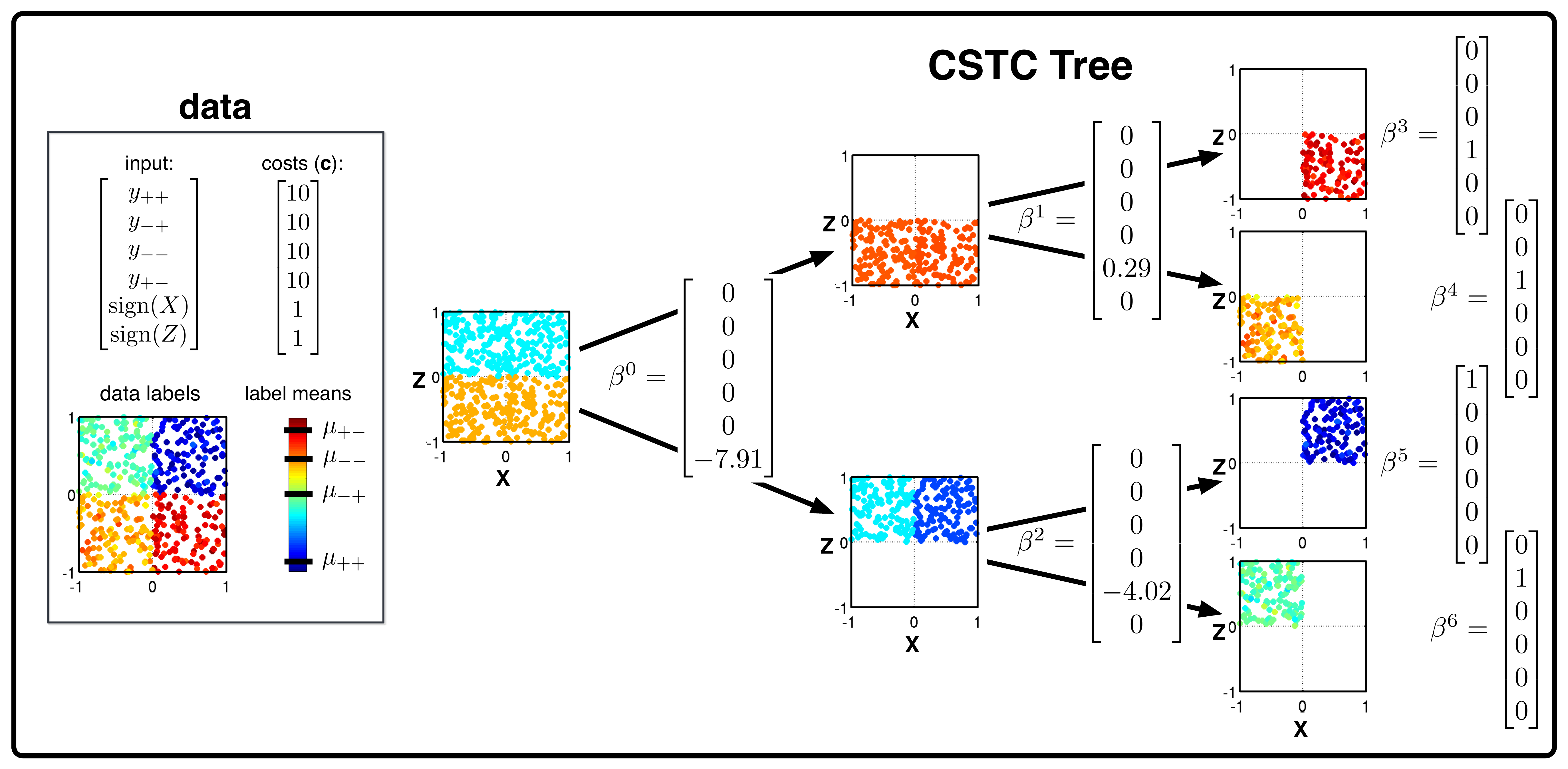}
}
\caption{CSTC on synthetic data. The box at left describes the artificial data set. The rest of the figure shows the CSTC tree built for the data set. At each node we show a plot of the predictions made by that classifier. After each node we show the weight vector that was selected to make predictions and send instances to child nodes (if applicable).}
\label{fig:split}
\end{figure*}

\subsection{Optimization Details}

There are many techniques to minimize the loss in (\ref{eq:nodeloss}). 
We use a cyclic optimization procedure, solving $\frac{\partial{\cal L}}{\partial (\bl^k, \theta^k)}$ for each classifier node $v^k$ one at a time, keeping all other nodes fixed.
For a given classifier node $v^k$, the traversal probabilities $p^j_i$ of a descendant node $v^j$ and the probability of an instance reaching a terminal element $p^l$ also depend on $\bl^k$ and $\theta^k$ (through its recursive definition)
and must be incorporated into the gradient computation. 

To minimize (\ref{eq:nodeloss}) with respect to parameters $\bl^k,\theta^k$, we use the lemma below to overcome the non-differentiability of the square-root terms (and $l_1$ norms) resulting from the 
$l_0$-relaxations (\ref{eq:l0relaxation}). 

\textbf{Lemma 1.} \emph{Given any $g(x) > 0$}, the following holds:
\begin{align}
	\sqrt{g(x)} = \min_{z>0}\frac{1}{2}\Bigg[\frac{g(x)}{z} + z\Bigg]. \label{eq:lemma1}
\end{align}
The lemma can be proved as $z = \sqrt{g(x)}$ minimizes the function on the right hand side. Further, it is shown in~\cite{boyd2004convex} that the right hand side is jointly convex in $x$ and $z$, so long as $g(x)$ is convex. 

For each square-root or $l_1$ term we introduce an auxiliary variable (i.e., $z$ above) and alternate between minimizing  the loss in (\ref{eq:nodeloss}) with respect to $\bl^k,\theta^k$ and the auxiliary variables. The former is performed with conjugate gradient descent and the latter can be computed efficiently in closed form. 
This pattern of block-coordinate descent followed by a closed form minimization is repeated until convergence. Note that the loss is guaranteed to converge to a fixed point because each iteration decreases the loss function, which  is bounded below by $0$. 

\textbf{Initialization.} The minimization of eq.~(\ref{eq:nodeloss}) is non-convex and therefore initialization dependent. However, minimizing eq.~(\ref{eq:nodeloss}) with respect to the parameters of leaf classifier nodes \emph{is convex}, as the loss function, after substitutions based on lemma 1, becomes jointly convex (because of the lack of descendant nodes). 
We therefore initialize the tree top-to-bottom, starting at $v^0$, and optimize over $\bl^k$ by minimizing (\ref{eq:nodeloss}) while considering all descendant nodes of $v^k$ as ``cut-off'' (thus pretending node $v^k$ is a leaf).

\textbf{Tree pruning.} To obtain a more compact model and to avoid overfitting, the CSTC tree can be pruned with the help of a validation set. As each node is a classifier, we can apply the CSTC tree on a validation set and compute the validation error at each node. We prune away nodes that, upon removal, do not decrease the performance of CSTC on the validation set (in the case of ranking data, we even can use validation NDCG as our pruning criterion).

\textbf{Fine-tuning.}
The relaxation in (\ref{eq:l0relaxation}) makes the exact $l_0$ cost terms differentiable and is well suited to approximate \emph{which} dimensions in a vector $\bl^k$ should be assigned non-zero weights. 
The mixed-norm does however impact the performance of the classifiers because (different from the $l_0$ norm)  larger weights in $\bl$ incur larger penalties in the loss. 
We therefore introduce a post-processing step to correct the classifiers from this unwanted regularization effect. 
We re-optimize all \emph{predictive} classifiers (classifiers with terminal element children, \emph{i.e.} classifiers that make final predictions), while clamping all features with zero-weight to strictly remain zero. 
\begin{align}
	\min_{\bar\bl^k} & \sum_{i} p_i^k\ell(\phi(\x_i)^\top\bar\bl^{k},y_i) + \rho |\bar\bl^{k}|\nonumber\\
	\textrm{ subject to: } & \bar\beta^k_t=0 \textrm{ if } \beta^k_t=0. \label{eq:fine-tuning}
\end{align}
The final CSTC tree uses these re-optimized weight vectors $\bar\bl^k$ for all predictive classifier nodes $v^k$.

\section{Results}
In this section, we first evaluate CSTC on a carefully constructed synthetic data set to test our hypothesis that CSTC learns specialized classifiers that rely on different feature subsets. We then evaluate the performance of CSTC on the large scale Yahoo! Learning to Rank Challenge data set and compare it with state-of-the-art algorithms. 

\subsection{Synthetic data} 
We construct a synthetic regression dataset, sampled from the four quadrants of the $X,Z$-plane, where $X\!=\!Z\!=\![-1,1]$. The features belong to two categories: cheap features, $sign(x), sign(z)$ with cost $c\!=\!1$, which can be used to identify the quadrant of an input; and four expensive features $y_{++},y_{+-},y_{-+},y_{--}$ with cost $c\!=\!10$, which represent the exact label of an input if it is from the corresponding region (a random number otherwise). \Matt{Since in this synthetic data set we do not transform the feature space, we have $\phi(\x)\!=\!\x$, and $\Fb$ (the weak learner feature-usage variable) is the $6\!\times\!6$ identity matrix.} By design, a perfect classifier can use the two cheap features to identify the sub-region of an instance and then extract the correct expensive feature to make a perfect prediction. The minimum feature cost of such a perfect classifier is exactly $c\!=\!12$ per instance. 
The labels are sampled from Gaussian distributions with quadrant-specific means $\mu_{++},\mu_{-+},\mu_{+-},\mu_{--}$ and variance $1$. 
Figure~\ref{fig:split} shows the CSTC tree and the predictions of test inputs made by each node. In every path along the tree, the first two classifiers split on the two cheap features and identify the correct sub-region of the input.  The final classifier extracts a single expensive feature to predict the labels. 
As such, the mean squared error of the training and testing data both approach 0.

\subsection{Yahoo! Learning to Rank} 
To evaluate the performance of CSTC on real-world tasks, we test our algorithm on the public Yahoo! \!Learning to Rank Challenge data set\footnote{\url{http://learningtorankchallenge.yahoo.com}}  \citep{chapelle2011yahoo}. The set contains 19,944 queries and 473,134 documents. Each query-document pair $\x_i$ consists of 519 features.
 An extraction cost, which takes on a value in the set $\{1,5,20,50,100,150,200\}$, is associated with each feature\footnote{The extraction costs were provided by a Yahoo! employee. }.  The unit of these values is the time required to evaluate a weak learner $h_t(\cdot)$.
The label $y_i \in \{4, 3, 2, 1, 0\}$ denotes the relevancy of a document to its corresponding query, with $4$ indicating a perfect match. 
In contrast to~\citet{chen2011}, we do not inflate the number of irrelevant documents (by counting them $10$ times). 
We measure the performance using NDCG@5 ~\citep{jarvelin2002cumulated}, a preferred ranking metric when multiple levels of relevance are available.  
Unless otherwise stated, we
 restrict CSTC to a maximum of $10$ nodes. 
All results are obtained on a desktop with two 6-core Intel i7 CPUs. Minimizing the global objective requires less than 3 hours to complete, and fine-tuning the classifiers takes about 10 minutes.

\begin{figure}[t]
\centerline{
\includegraphics[width = 1\columnwidth]{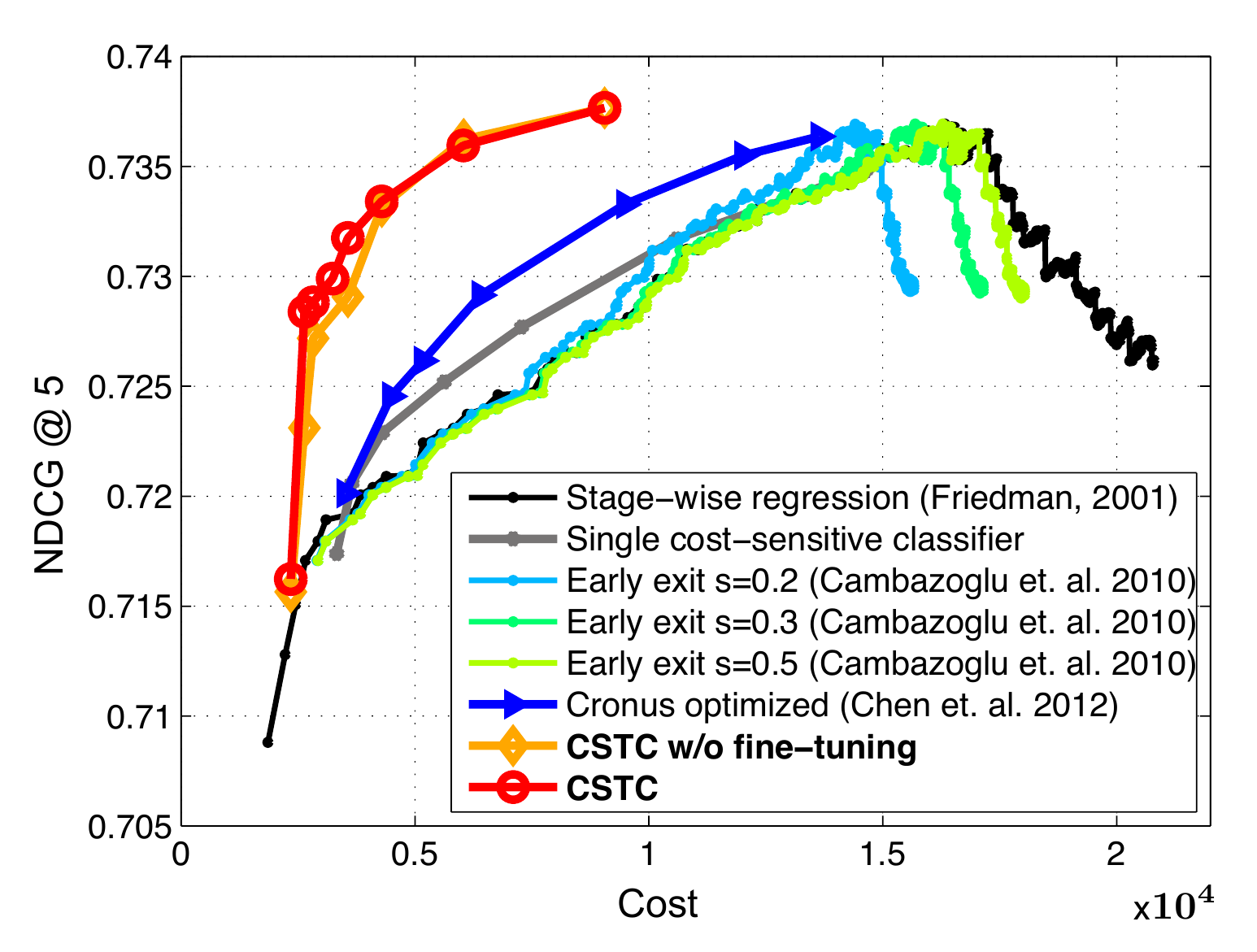}
}
\caption{The test ranking accuracy (NDCG@5) and cost of various cost-sensitive classifiers. CSTC maintains its high retrieval accuracy significantly longer as the cost-budget is reduced. Note that fine-tuning does not improve NDCG significantly because, as a metric, it is \emph{insensitive} to mean squared error.
}
\label{fig:versus}
\end{figure}

\begin{figure}[t]
\begin{center}
\includegraphics[width = \columnwidth]{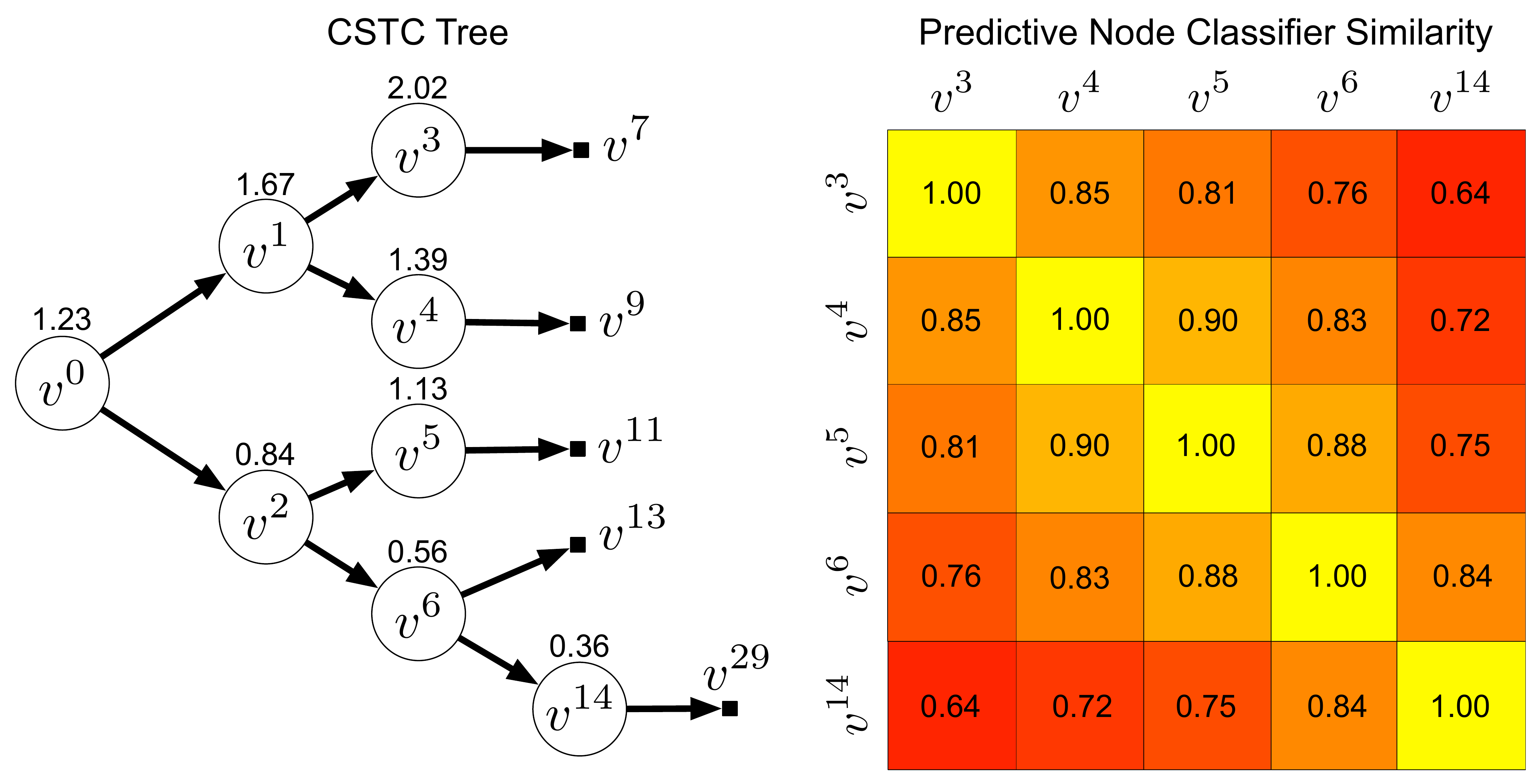}
\end{center}
\caption{
(\emph{Left}) The pruned CSTC-tree generated from the Yahoo! Learning to Rank data set. (\emph{Right}) Jaccard similarity coefficient between classifiers within the learned CSTC tree.
\label{fig:jaccard}
}
\end{figure}

\textbf{Comparison with prior work.}
Figure~\ref{fig:versus} shows a comparison of CSTC with several recent algorithms for test-time cost-sensitive learning. We show NDCG versus cost (in units of weak learner evaluations). The plot shows different stages in our derivation of CSTC: the initial cost-insensitive ensemble classifier 
$H'(\cdot)$~\citep{friedman2001greedy} from section~\ref{sec:method} (\emph{stage-wise regression}), a \emph{single cost-sensitive classifier} as described in eq.~(\ref{eq:onenode}), 
the CSTC tree (\ref{eq:nodeloss}) and CSTC tree \emph{with} fine-tuning (\ref{eq:fine-tuning}).
We obtain the curves by varying the accuracy/cost trade-off parameter $\lambda$ (and perform early stopping based on the validation data, for fine-tuning). 
For CSTC tree we evaluate six settings, $\lambda \!=\! \{ \frac{1}{3}, \frac{1}{2}, 1,2,3,4,5,6 \}$. 
In the case of \emph{stage-wise regression}, which is not cost-sensitive, the curve is simply a function of boosting iterations.

For competing algorithms, we include \emph{Early exit} \citep{cambazoglu2010early} which improves upon stage-wise regression by short-circuiting the evaluation of unpromising documents at test-time, reducing the overall test-time cost.  The authors propose several criteria for rejecting inputs early and we use the best-performing method ``early exits using proximity threshold".  For \emph{Cronus}~\citep{chen2011}, we use a cascade with a  maximum of 10 nodes. All hyper-parameters (cascade length, keep ratio, discount, early-stopping) were set based on a validation set. The cost/accuracy curve was generated by varying the corresponding trade-off parameter, $\lambda$.

As shown in the graph, CSTC significantly improves the cost/accuracy trade-off curve over all other algorithms. 
The power of Early exit is limited in this case as the test-time cost is dominated by feature extraction, rather than the evaluation cost.  
Compared with Cronus, CSTC has the ability to identify features that are most beneficial to different groups of inputs. It is this ability, which allows CSTC to maintain the high NDCG significantly longer as the cost-budget is reduced. 

Note that CSTC with fine-tuning only achieves very tiny improvement over CSTC without it. Although the fine-tuning step decreases the mean squared error on the test-set, it has little effect on NDCG, which is only based on the relative ranking of the documents (as opposed to their exact predictions). Moreover, because we fine-tune prediction nodes until validation NDCG decreases, for the majority of $\lambda$ values, only a small amount of fine-tuning occurs. 

\textbf{Input space partition.}
Figure~\ref{fig:jaccard} (\emph{left}) shows a pruned CSTC tree ($\lambda\!=\! 4$) for the Yahoo! data set. The number above each node indicates the average label of theWg inputs passing through that node. We can observe that different branches aim at different parts of the input domain. In general, the upper branches focus on correctly classifying higher ranked documents, while the lower branches target low-rank documents. 
Figure~\ref{fig:jaccard} (\emph{right}) shows the Jaccard matrix of the predictive classifiers  $(v^3, v^4, v^5, v^6, v^{14})$ from the same CSTC tree. 
The matrix shows a clear trend that the Jaccard coefficients decrease monotonically away from the diagonal. This indicates that classifiers share fewer features in common if their average labels 
are further apart---the most different classifiers $v^3$ and $v^{14}$ have only $64\%$ of their features in common---and validates that classifiers in the CSTC tree extract different features in different regions of the tree. 

\begin{figure}[t]
\centerline{
\includegraphics[width = 1\columnwidth]{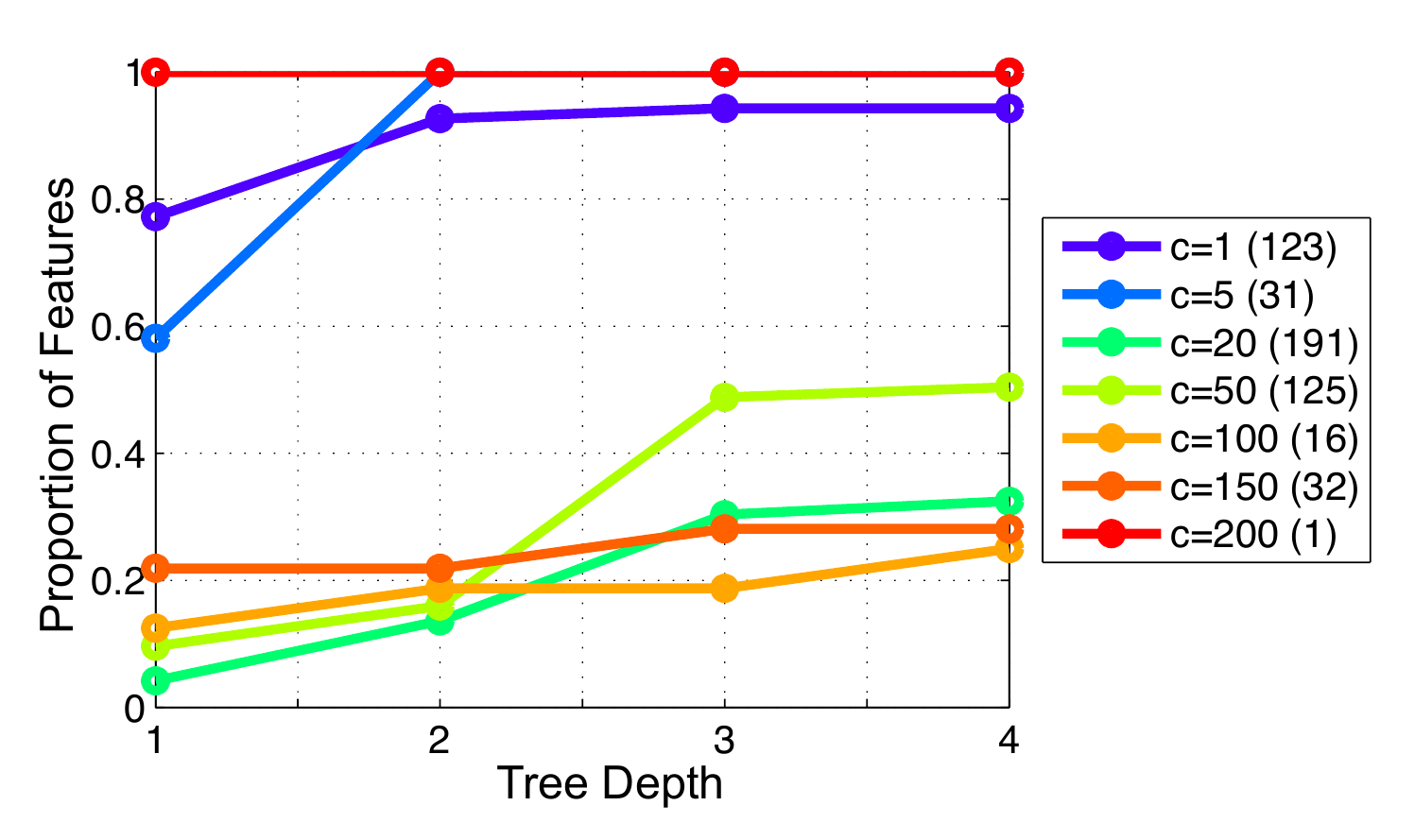}
}
\caption{The ratio of features, grouped by cost, that are extracted at different depths of CSTC (the number of features in each cost group
is indicated in parentheses in the legend). More expensive features ($c \ge 20$) are gradually extracted as we go deeper.}
\label{fig:features}
\end{figure}

\textbf{Feature extraction.}
\label{results:feats}
We also investigate the features extracted in individual classifier nodes.  Figure~\ref{fig:features} shows the fraction of features, with a particular cost, extracted at different depths of the CSTC tree for the Yahoo! data. We observe a general trend that as depth increases, more features are being used. However, cheap features ($c\le 5$) are fully extracted early-on, whereas expensive features ($c\ge 20$) are extracted by classifiers sitting deeper in the tree, where each individual classifier only copes with a small subset of inputs. The expensive features are used to classify these subsets of inputs more precisely. The only feature that has cost $200$ is extracted at all depths---which seems essential to obtain high NDCG~\cite{chen2011}.

\section{Conclusions}
We introduce Cost-Sensitive Tree of Classifiers (CSTC), a novel learning algorithm that explicitly addresses the trade-off between accuracy and expected test-time CPU cost in a principled fashion. The CSTC tree partitions the input space into sub-regions and identifies the most cost-effective features for each one of these regions---allowing it to match the high accuracy of the state-of-the-art at a small fraction of the  cost. We obtain the CSTC algorithm by formulating the expected test-time cost of an instance passing through a tree of classifiers and relax it into a continuous cost function. This cost function can be minimized while learning the parameters of all classifiers in the tree jointly. 
By making the test-time cost vs. accuracy tradeoff explicit  we enable high performance classifiers that fit into computational budgets and can reduce unnecessary energy consumption in large-scale industrial applications. Further,  engineers can design highly specialized features for particular edges-cases of their input domain and CSTC will automatically incorporate them on-demand into its tree structure.

\footnotesize
\paragraph{Acknowledgements}
KQW, ZX, MK, and MC are supported by NIH grant U01 1U01NS073457-01 and NSF grants 1149882 and 1137211. The authors thank John P. Cunningham for clarifying discussions and suggestions. 

\bibliography{tronus}
\bibliographystyle{icml2012}

\end{document}